\documentclass[letterpaper, 10 pt, conference]{ieeeconf}  

\IEEEoverridecommandlockouts                              

\usepackage{graphics} 
\usepackage{epsfig} 
\usepackage{mathptmx} 
\usepackage{times} 
\usepackage{amsmath} 
\usepackage{amssymb}  
\usepackage{subfigure} 
\usepackage{xcolor}
\usepackage{stfloats}
\usepackage{booktabs}
\usepackage{hyperref}

\title{\LARGE \bf
Contact-Rich and Deformable Foot Modeling for Locomotion Control of the Human Musculoskeletal System
}

\author{Haixin Gong$^{1}$, Chen Zhang$^{1}$, Yanan Sui$^{1}$
\thanks{IEEE-RAS 24th International Conference on Humanoid Robots (Humanoids 2025). $^{1}$Tsinghua University, Beijing, China. This work is funded by STI2030-Major Projects-2022ZD0209400, NSFC-62206152, and Tsinghua Dushi Fund. Correspondence to Yanan Sui: {\tt\small ysui@tsinghua.edu.cn}}
}
\begin{document}

\maketitle
\thispagestyle{empty}
\pagestyle{empty}

\begin{abstract}

The human foot serves as the critical interface between the body and environment during locomotion.
Existing musculoskeletal models typically oversimplify foot-ground contact mechanics, limiting their ability to accurately simulate human gait dynamics. 
We developed a novel contact-rich and deformable model of the human foot integrated within a complete musculoskeletal system that captures the complex biomechanical interactions during walking.
To overcome the control challenges inherent in modeling multi-point contacts and deformable material, we developed a two-stage policy training strategy to learn natural walking patterns for this interface-enhanced model. 
Comparative analysis between our approach and conventional rigid musculoskeletal models demonstrated improvements in kinematic, kinetic, and gait stability metrics. Validation against human subject data confirmed that our simulation closely reproduced real-world biomechanical measurements. 
This work advances contact-rich interface modeling for human musculoskeletal systems and establishes a robust framework that can be extended to humanoid robotics applications requiring precise foot-ground interaction control.

\end{abstract}

\section{INTRODUCTION}
\label{sec:intro}

The human foot is the primary contact interface for interaction with the ground. More than just a supporting structure bearing the human body load, it also plays a crucial role in dynamically distributing forces across multiple contact points on the plantar surface and effectively absorbing impacts during movements. This facilitates stability and agility to interact with diverse environments. The contact-rich and deformable properties of the foot are essential for these functions, enabling increased contact area, reduced localized pressure peaks \cite{piazza2024analytical}, smoothened force transition \cite{farris2020foot}, and enhanced energy efficiency \cite{ker1987spring}, while also helping prevent injury \cite{van2021foot}.
Although recent advances in musculoskeletal modeling demonstrated improved structural and biomechanical precision, the deformable interface between the foot and ground is still lacking 
\cite{arnold2010model}, \cite{rajagopal2016full}, \cite{zuo2024self}, \cite{d2024dynamic}. Rigid body modeling of the foot in existing musculoskeletal models simplified the interface with very limited contact points, resulting in discrete but overly concentrated force distribution and insufficient shock absorption. Incorporating the contact-rich interface of the foot is therefore important for better modeling human-ground interaction and simulating realistic human locomotion via the human musculoskeletal system.
Furthermore, the control strategy, force distribution, and dynamic properties derived from more realistic human locomotion simulation could provide valuable insights into gait dynamics and biomechanics and facilitate humanoid robot design by bridging the sim-to-real gap.

\begin{figure}[tbp]
    \centering
    \includegraphics[width=0.95\linewidth]{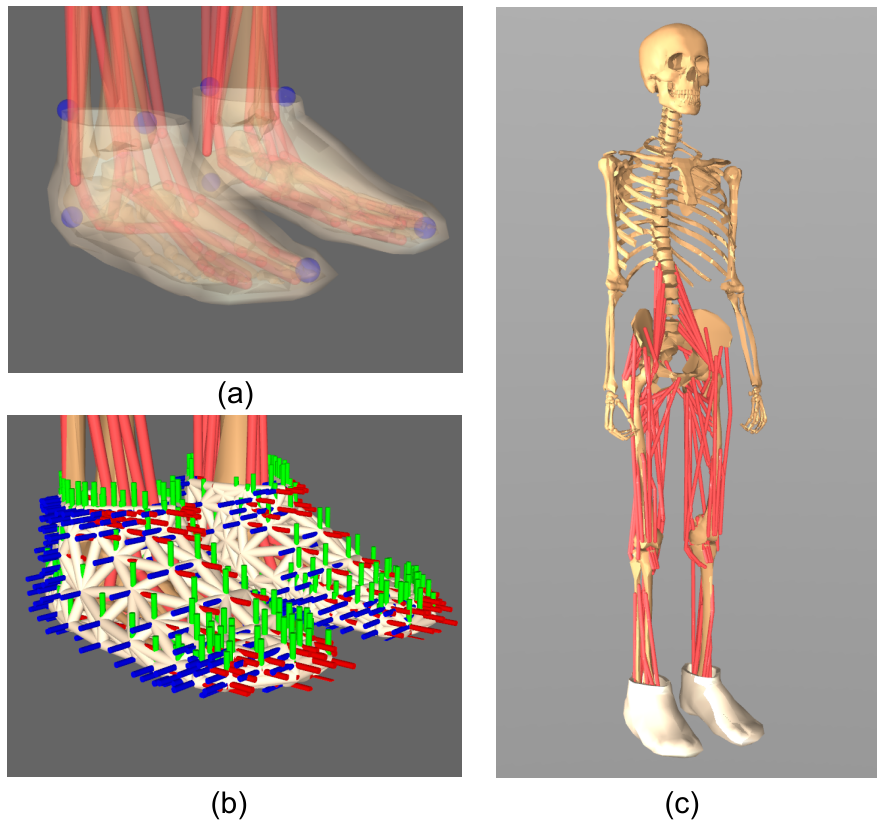}
    \caption{Contact-rich and deformable foot modeling. (a) Modeled feet with deformable materials. Semi-transparent meshes indicate the soft material and blue spheres indicate the attachment points to the rigid musculoskeletal model. (b) Contact-rich edges (off-white lines) and the coordinates (colored in red, blue, and green for the three axes) of vertices. (c) Adding the new feet to the musculoskeletal model \href{https://lnsgroup.cc/research/MS-Human}{MS-Human-700} to get the interface-enhanced model with bones, muscle-tendon units, and contact-rich feet.}
    \label{fig:foot_model}
\end{figure}

\begin{figure*}[tbp]
    \centering
    \includegraphics[width=0.95\linewidth]{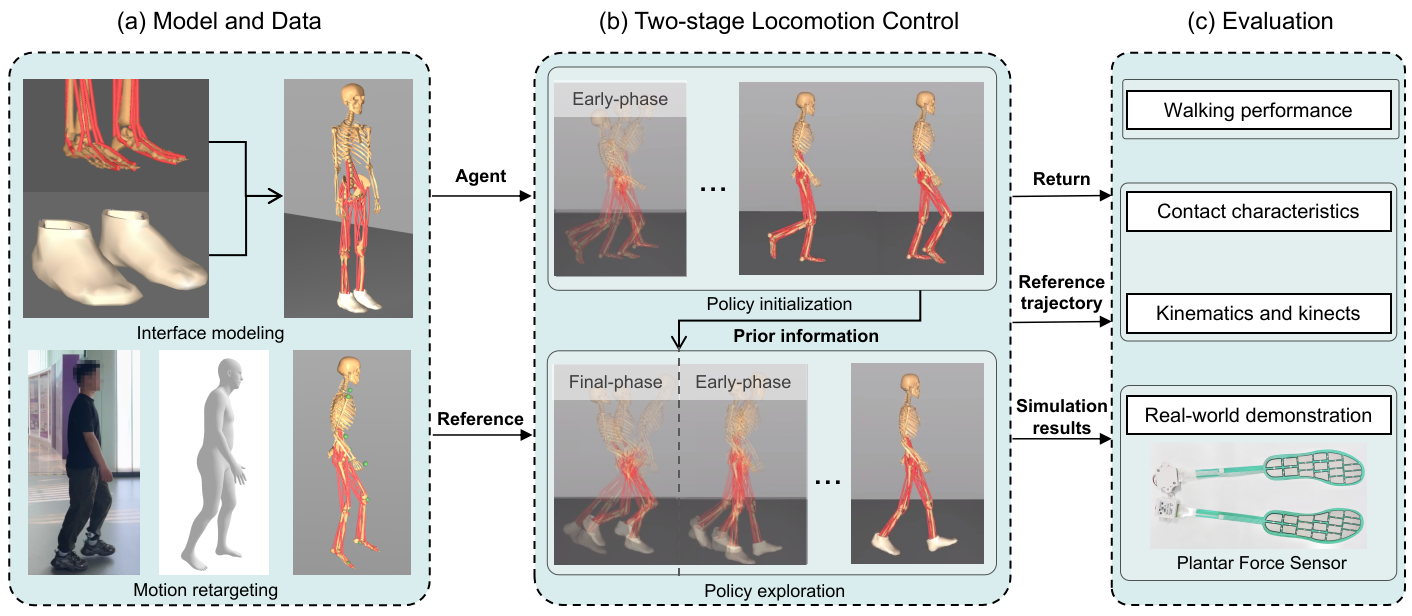}
    \caption{Overview of the framework for interface modeling and locomotion control. (a) Environment Preparation, including the musculoskeletal model integrated with enhanced foot interface modeling, motion retargeting from human behavior, and plantar force measurement for the human participant; 
    (b) Two-stage Locomotion Control, involving policy initialization on rigid model and subsequent policy exploration guided by prior knowledge and reference trajectories; 
    (c) Model Evaluation, including evaluations on walking performance, contact characteristics, and kinematic and kinetic fidelity of the simulation results compared to real-world data.}
    \label{fig:whole_pipeline}
\end{figure*}

Despite its potential to improve stability and adaptability during locomotion \cite{de2017design}, high-fidelity and contact-rich interface modeling of the human foot introduces significant challenges. 
Modeling the deformable properties of the human foot requires the integration of anatomical and biomechanical knowledge with feasible simulation solutions. 
The simulation stability would be degraded by complex contact and collision dynamics. This degradation could become more significant with increased deformable degrees of freedom (DoFs) and more complicated geometry. The finite element method (FEM) is often used to handle the soft body modeling problem for static simulation \cite{akrami2018subject}, \cite{kamal2024combined}, but it is computationally intensive and poorly suited for simulating tasks requiring complex dynamics like human locomotion \cite{korber2021comparing}. A variety of graphical modeling studies treated the foot as a simple structure such as a box or capsule consisting of one or two bodies for physical simulation \cite{park2020multi}, lacking structural complexity needed to model realistic and adaptive behaviors.
Thus, a novel modeling approach is needed to take into account the structural fidelity, deformable property, and contact-rich characteristics of the human foot while ensuring computational efficiency and simulation stability when integrated into a musculoskeletal system.

Deep reinforcement learning methods represent the state-of-the-art in controlling the human musculoskeletal system for their flexibility and adaptability in generating complex movements. However, the musculoskeletal system is high-dimensional and overactuated, posing challenges in realistic and biomechanically plausible control \cite{weimotion}.
Moreover, incorporating the contact-rich foot as a deformable interface between the musculoskeletal system and the ground introduces more DoFs and model parameters, further increasing the complexity of learning and control for realistic locomotion. Therefore, to achieve high performance in human walking simulation by such an interface-enhanced musculoskeletal model, we proposed an efficient policy training strategy with a two-stage imitation learning (IL) framework. Furthermore, we conducted real-world experiments to capture human walking sequences and plantar pressure sensor data simultaneously. Kinematic and kinetic comparisons were conducted between the simulation results and the real-world measurements for model evaluations.
The main contributions of this paper are: 
(1) Design of a contact-rich and deformable foot model integrated with the human musculoskeletal system, capable of reproducing complex material properties, structural mechanics, and multi-point contact dynamics.
(2) Development of a two-stage learning strategy that effectively addresses the computational challenges of controlling high-dimensional musculoskeletal systems with deformable contact interfaces.
(3) Evaluation through human experiments, demonstrating the benefits of incorporating the contact-rich foot and validating the transferability of the interface-enhanced musculoskeletal model.


\section{RELATED WORK}
\label{sec:related_work}

Musculoskeletal models comprising bones, joints, and muscles-tendon units, simulated using physics engines on platforms like OpenSim \cite{delp2007opensim} and MuJoCo \cite{Mujoco}, are essential tools for computational analysis and simulation of human-like behaviors. These models typically implemented the foot as a kinematic chain of rigid segments connected by torque-producing actuators, lacking a deformable interface representation. This simplification limited their ability to capture the complex biomechanical properties of the plantar surface. While several studies aimed to improve simulation accuracy by modeling the ankle-foot complex within the musculoskeletal system, they largely remained constrained to rigid foot representations and neglected the deformable nature of the foot-ground interaction \cite{d2024dynamic}, \cite{maharaj2022modelling}, \cite{sikidar2022open}. Well-developed programming interfaces for reinforcement learning, such as LocoMujoco \cite{locomujoco} and Gymnasium \cite{towers2024gymnasium}, facilitate the development of locomotion control algorithms for whole-body musculoskeletal models \cite{weimotion}, \cite{he2024dynsyn}. However, the reliance on over-simplified models could constraint the attainable upper bound of performance. A soft-body foot design integrated with the skeletal model was recently proposed for walking simulation \cite{loke2024personalised}, representing a promising advancement in interface modeling. However, in addition to lacking muscle-tendon units, the model was over-simplified in geometry for ease of control and was unable to capture the contact dynamics. Integrating a more detailed foot model into the musculoskeletal system and validating its controllability and compliance would be an important step toward more reliable and realistic locomotion control. 

In robotics and prosthetics, foot designs have largely remained rigid and flat to simplify actuation, sensing, and control. As a result, the efforts to achieve stable locomotion often shifted to the upper body and joints, limiting the adaptability and efficiency of the systems. Soft and deformable foot designs have been proposed to improve ground compliance and adaptability for quadrupedal robots \cite{shepherd2011multigait}, offering new perspectives for advancing humanoid locomotion. While some studies explored interface modeling to improve the adaptability of humanoid robots on uneven terrain \cite{piazza2024analytical}, \cite{park2020multi}, \cite{colasanto2015bio}, these designs typically employed multi-segmented feet with articulated heel, midfoot, and toe sections. Although such configurations approximated the structural flexibility of the human foot to some extent, they still fell short of replicating the inherent compliance and deformation. \cite{de2017design} proposed a design pipeline for deformable soles in humanoid robots, demonstrating the importance of foot shape in locomotion performance, even though the design was not based on real human feet. Developing a human-like compliant interface both geometrically and physically is crucial for enhancing locomotion performance in terms of efficiency, stability, and adaptability.

Despite that the rigid foot remains a standard component in most robotic designs, deformable foot modeling has gained considerable attention in biomechanics. In this domain, FEM is the dominant approach for modeling soft body with complex geometry under diverse loading and boundary conditions. Teran et al. \cite{teran2005robust} pioneered the use of FEM for real-time tissue simulation, showing the importance of nonlinear elasticity in capturing complex deformations. Amounts of detailed FEM-based models were developed for the study of plantar pressure distribution \cite{akrami2018subject, bruening2024new}, plantar deformation \cite{filardi2018finite}, and structural effects \cite{mancera2025quantitative} of the foot. 
Although the primary focus of these models was on reproducing and analyzing human foot kinematics and kinetics \cite{park2020multi}—which restricted their applicability to static or quasi-static simulations and hindered efficient scaling for dynamic, real-time scenarios such as locomotion control—they also incorporated CT or MRI imaging as anatomical references and applied soft material properties and constraints to represent soft components \cite{akrami2018subject}, \cite{mancera2025quantitative}, \cite{mitsuhashi2009bodyparts3d}, \cite{kwan2010effect}. This modeling approach can provide valuable prior knowledge of foot morphology, informing human-inspired foot design.

\section{METHODS}
\label{sec:methods}

\subsection{Contact-rich Foot Modeling} 
\label{sec:musculoskeletal_model_and_simulation}

\subsubsection{Human Foot Model}
\label{sec:human_foot_model}

We used the anatomical model from BodyParts3D \cite{mitsuhashi2009bodyparts3d} for reference, thereby enabling mesh-based geometric modeling to initialize the interface design with vertices, edges and mesh structures. The ankle-foot complex was then extracted using the animation software Blender. For convenience, the left foot model was built symmetrically from the right foot. To balance simulation stability and computational cost, we simplified the mesh of the foot. The resulting foot interface model contains 223 vertices and 426 triangular elements per foot, as shown in Fig.~\ref{fig:foot_model}(b). This deformable mesh was then integrated into the musculoskeletal system and served as the contact interface for foot-ground interaction. As our focus was primarily on the locomotion over the human-ground interface, the upper limb musculature was omitted to reduce model complexity. The lower extremity model was derived from the whole-body musculoskeletal model MS-Human-700 \cite{zuo2024self}, comprising 100 muscle-tendon actuators and 36 active DoFs. Through appropriate geometric simplification and scaling, we constructed a physically plausible foot interface model with bounding box dimensions of 29.8~$cm$ $\times$ 11.6~$cm$ $\times$ 12~$cm$.

\begin{table}[tbp]\normalsize
    \caption{Mechanical properties of the flex element}
    \label{tab:flex_element_properties}
    \begin{center}
    \begin{tabular}{ccc}
    \toprule
    \textbf{Property} & \textbf{Value} & \textbf{Unit} \\
    \midrule
    Radius & 0.005 & m \\
    Mass & 0.05 & kg \\
    Poisson's ratio & 0.49 & - \\
    Young's modulus & 1.0e5 & Pa \\
    Acceleration reference & (-5e4 -1e3) & - \\
    Constraint impedance  & (0.1 0.9 0.001 0.5 2) & - \\
    \bottomrule
\end{tabular}
\end{center}
\end{table}

\subsubsection{Interface Modeling}
\label{sec:interface_modeling}

Interface modeling is critical for representing interactions between the musculoskeletal system and the environment. At the foot–musculoskeletal interface, the deformable foot geometry was linked to the calcaneus bone as a sub-body of the corresponding body element and implemented as a child node in the kinematic tree. To preserve anatomical attachment, the boundary vertices of the interface mesh were pinned by removing all associated DoFs. Since the physical engine permits interpenetration between bodies, equality constraints were explicitly imposed to connect vertices experiencing frequent impacts capturing realistic biomechanical behavior, as shown in Fig.~\ref{fig:foot_model}(a). These constraints emulate the compliant attachment between the foot surface and the musculoskeletal structure in humans, allowing limited relative motion between the soft tissue and the rigid bones \cite{nester2007foot}.

At the foot–ground interface, the flex element modeling approach was employed, which automatically generates a set of deformable moving bodies with corresponding joints based on the mesh vertices. By default, each vertex is assigned three translational DoFs, resulting in high computational costs. Therefore, to reduce computational load and improve numerical stability, each vertex was restricted to a single radial translational DoF. Additionally, equality constraints were applied to the flex edges to preserve their initial lengths. The DoFs introduced by the flex elements were treated as passive DoFs, allowing the foot to deform naturally in response to contact dynamics rather than being directly controlled by the policy. This separation facilitates more realistic gait behaviors.

\subsubsection{Contact Properties}
The physical parameters provide a preliminary characterization of the deformable properties of the interface model. In practice, contact solver parameters play a critical role in capturing contact dynamics and ensuring simulation stability. We experimented with plugin-based, elasticity-based, and edge-based modeling approaches, and found that the edge-based method offers greater robustness in simulating the interface-enhanced musculoskeletal system.
A variety of attributes defining the behavior of flex elements were determined based on a literature review \cite{akrami2018subject}, \cite{mancera2025quantitative}, \cite{kwan2010effect}. Additionally, carefully designed solver parameters—namely constraint impedance and reference acceleration—further improve contact realism and compliance. All relevant physical and solver parameters were summarized in Table~\ref{tab:flex_element_properties}. Notably, we adopted the negative format for the acceleration reference, which is advantageous for flexible interface modeling.

\subsection{Locomotion Control}
Controlling the interface-enhanced musculoskeletal model presents significant challenges due to its high-dimensional state space and non-linear contact dynamics. Training a policy from scratch was proved inadequate for this task. Given the similar kinematic structure and task-specificity between the rigid and interface-enhanced models, we introduced a two-stage policy training method to generate natural walking patterns. Below we detailed the pipeline of data preparation and the two-stage training strategy.

\subsubsection{Motion Retargeting}
We recorded videos of a subject performing walking tasks at a normal speed, as shown in Fig.~\ref{fig:whole_pipeline}(a), and used WHAM \cite{WHAM} to generate SMPL parameters \cite{SMPL}, which provide a parametric representation of human motion including pose and shape. To convert human poses to the musculoskeletal model, we defined $N=18$ hand-crafted sites $M \in \mathbb{R}^{3N}$ representing correspondences between SMPL body keypoints and musculoskeletal joints. The bone length was then scaled to real human proportions based on the site-to-site distance. Consequently, the joint kinematics are obtained by solving the following optimization problem, where the objective function is formulated as:

\begin{equation}
    \min_{q} \sum_{i=1}^{N} \| p_i(q) - p_i \|^2 + \lambda \|q\|^2
\end{equation}

where $p_i$ denotes the position of $i$-th site, $q \in \mathbb{R}^{33}$ represents the optimized joint angles excluding global translations, $\lambda$ is a regularization parameter. Note that as introduced in Section.~\ref{sec:musculoskeletal_model_and_simulation}, the DoFs introduced by the flex elements are considered as passive. Therefore, the dimensionality of the variable $q$ remains consistent between the interface-enhanced musculoskeletal model and the rigid model. The damped least squares method \cite{buss2004introduction} was employed in the optimization to avoid singularity issues. The optimized results are illustrated in Fig.~\ref{fig:whole_pipeline}(a), showing both the target keypoints (green spheres) and the optimized poses of the musculoskeletal model performing the same motion. 

After retargeting the body keypoints to the musculoskeletal model, the resulting joint trajectories were processed to generate suitable expert demonstrations for policy training. Joint velocities were computed using finite difference methods and combined with joint positions to form part of the complete state representation required by the environment. Following interpolation and filtering, the collected data were temporally aligned to match the frequency of 100~$Hz$.

\subsubsection{Reinforcement Learning for Walking}
\label{sec:reinforcement_learning_for_walking_locomotion}
For locomotion control, the DoFs and muscle–tendon actuators corresponding to the upper body were disabled both in the rigid model and the interface-enhanced musculoskeletal model. To ensure comparability of experimental results, the observation and action spaces were kept consistent across the two environments. A mask was applied to exclude the joint kinematics of the flex body from the observation space, and the reward function was defined based on this masked observation. DynSyn \cite{he2024dynsyn} was employed as the foundational policy owing to its proven effectiveness in controlling musculoskeletal humanoids for locomotion tasks.

In the first stage, we trained the initial policy on the rigid musculoskeletal model using the reference trajectory. This stage is computationally efficient due to the simplified contact dynamics. In the second stage, we adapted the initial policy and continued training on the interface-enhanced musculoskeletal model, encouraging the agent to mimic the kinematics of the rigid model. Specifically, this prior knowledge guided the agent toward beneficial regions of the state space, as intuitively illustrated in Fig.~\ref{fig:whole_pipeline}(b). During the early phase of training, the initialized agent exhibited falling behaviors similar to those of the rigid model, suggesting a more constrained and structured exploration. In contrast, the agent trained from scratch experienced more irregular behaviors due to the intricate contact dynamics, resulting in greater variability despite eventual convergence.

Note that similar kinematic performance does not imply identical dynamics, as the designed reward function primarily targets the kinematic distribution in the observation space and lacks direct sensitivity to complex contact dynamics and kinetic behaviors. Therefore, further analysis of walking patterns is essential to reveal the physical and biomechanical distinctions between the two models under the same walking trajectories.

\section{Experiments}

The experiments were designed to evaluate the controllability as well as the physical and biomechanical realism of the proposed interface-enhanced musculoskeletal model, and consisted of three parts:
(1) Training and Evaluation: to validate the effectiveness of the proposed policy training strategy.
(2) Stability and Realism Validation: to analyze and compare the walking patterns of the two models through contact characteristics and kinetic analysis.
(3) Real-world Demonstration: to preliminarily assess the realism and transferability of the proposed model by comparing simulation results with real human data.

\begin{figure*}[tbp]
    \centering
    \includegraphics[width=0.9\linewidth]{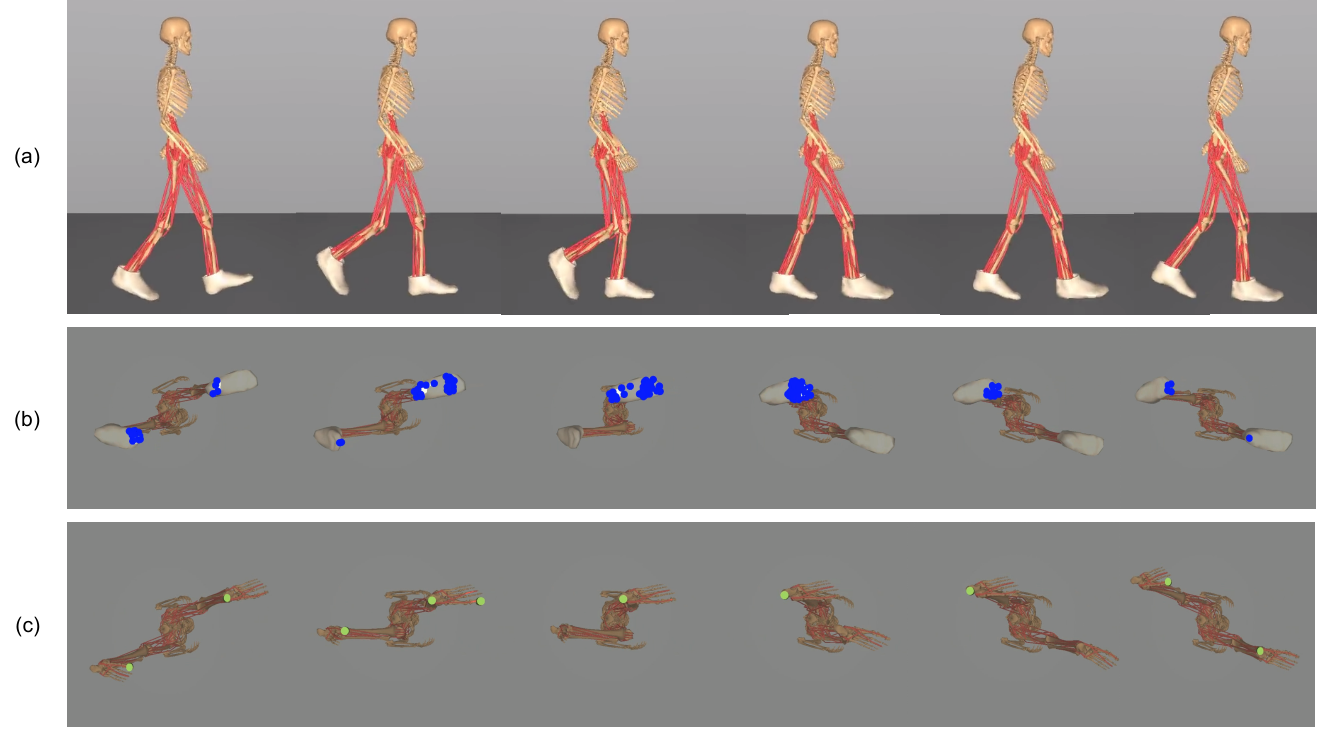}
    \caption{Locomotion sequence by the interface-enhanced musculoskeletal model. (a) Key frames of the learned walking behavior. (b) Contact points of the contact-rich foot from the bottom-up view, represented as blue dots and corresponding to the time points of the walking sequence in (a).
    (c) Contact points of the rigid foot from the bottom-up view, represented as green dots.}
    \label{fig:walking_sequence}
\end{figure*}

\subsection{Training and Evaluation}
To address the high-dimensional, contact-rich nature of the interface-enhanced musculoskeletal model, we adopted a two-stage training strategy. The environment was configured with a control frequency of 50~$Hz$ and a simulation frequency of 500~$Hz$. The observation space comprised joint positions, velocities, accelerations, muscle activations, actuator forces, and center-of-mass information, totaling 552 dimensions. The action space consisted of 100 muscle activation signals for lower-body control. The reward function was designed to encourage stable locomotion while maintaining natural gait patterns, using a weighted combination of multiple objectives formulated as:

\begin{equation}
    r = w_{q} r_{q} + w_{{\dot{q}}} r_{\dot{q}} + w_{act} r_{act} + w_{vel} r_{vel} + w_{healthy} r_{healthy}
\end{equation}

The joint position tracking term $r_{q}$ (weight: 50) penalizes deviations from the reference trajectory. The joint velocity tracking term $r_{\dot{q}}$ (weight: 0.1) minimizes velocity discrepancies relative to the target trajectory. The muscle activation penalty $r_{act}$ (weight: 1) promotes energy efficiency by discouraging excessive muscle activations. The forward velocity tracking term $r_{vel}$ (weight: 5) encourages the agent to maintain the target walking speed of 1.25~$m/s$. Finally, the posture stability term $r_{healthy}$ (weight: 100) ensures a stable upright posture throughout locomotion, preventing falls and promoting biomechanically realistic movements.

Both training stages employed the SAC algorithm \cite{pmlr-v80-haarnoja18b} with the DynSyn policy architecture \cite{he2024dynsyn}, using a learning rate of 5$\times$10$^{-4}$ with linear decay scheduling. The neural network architecture consisted of separate policy and Q-function networks, each with hidden layers of sizes [2048, 1024, 1024]. The policy incorporated 20 hierarchical action groups to capture the complex coordination patterns between muscle groups and the flexible foot elements. As a result, stable walking locomotion was successfully achieved despite the complex contact properties and dynamics, as illustrated in Figure.~\ref{fig:walking_sequence}(a). 



\subsection{Stability and Realism Validation}
As introduced in Section~\ref{sec:reinforcement_learning_for_walking_locomotion}, despite both models achieved comparable kinematic performance, this does not indicate equivalence in their underlying locomotion dynamics (see Figure.~\ref{fig:walking_sequence}(b)(c)). In this case, we further looked into the contact characteristics and kinetic behaviors of the interface-enhanced model. Unless otherwise specified, all statistical results are reported as Mean ± Standard Error (SE), aggregated over 12 episodes comprising more than 70 gait cycles in total.

\subsubsection{Contact Characteristics}
\label{sec:contact_characteristics}

\begin{figure}[tbp]
    \centering
    \includegraphics[width=1\linewidth]{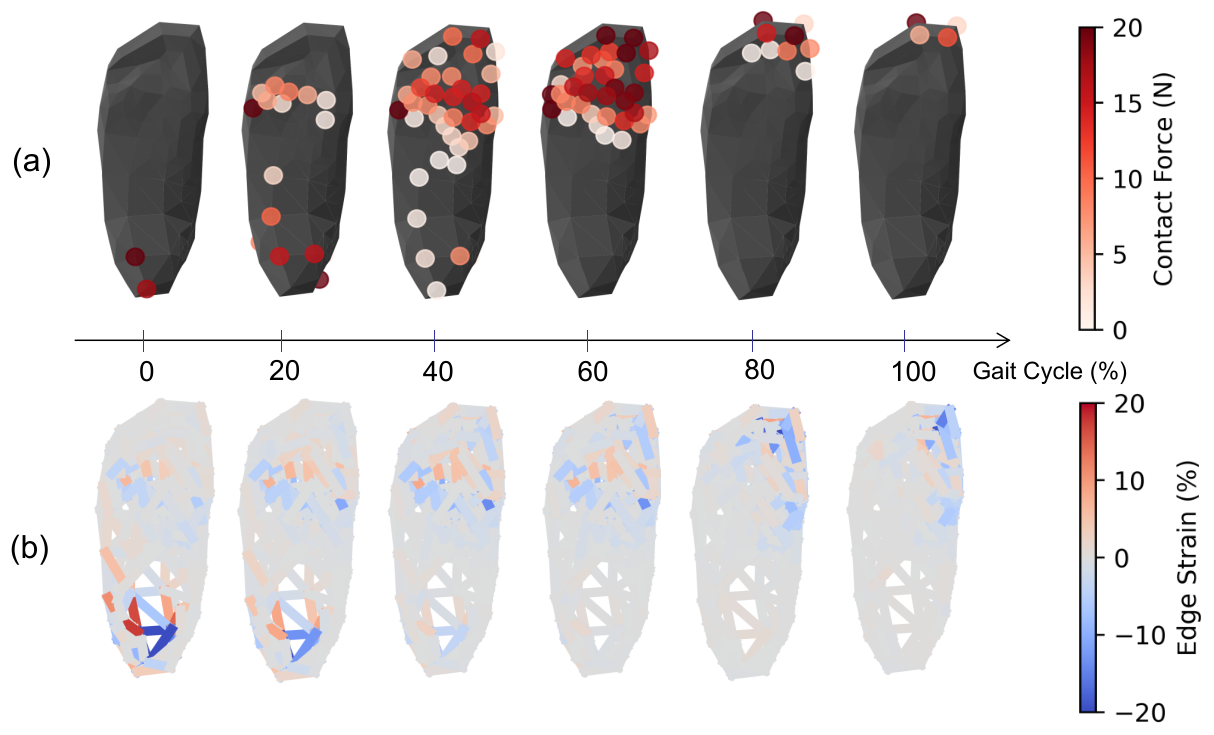}
    \caption{Considering the bilateral symmetry of the musculoskeletal model, only the right foot contact data are shown.
(a) Spatiotemporal distribution of plantar contact forces over a typical gait cycle, visualized as colored contact points, where color intensity encodes the contact force magnitude.
(b) Corresponding mesh deformation of the interface-enhanced model, represented as edge-wise displacements. Displacements are expressed as strains, with red indicating tensile deformation (stretching) and blue indicating compressive deformation (shrinking).}
    \label{fig:contact_points_and_foot_force}
\end{figure}

Benefit from the deformable foot, the interface-enhanced musculoskeletal model processes larger contact area and exhibits more contact dynamics, as illustrated in Figure.~\ref{fig:contact_points_and_foot_force}. Comparing the rigid model which has fewer contact points (see Figure.~\ref{fig:walking_sequence}(c)), the contact force map (see Figure.~\ref{fig:contact_points_and_foot_force}(a)) for the interface-enhanced model could reveal a more realistic progression of contact regions across the plantar surface as the gait cycle advances. The progression starts with heel strike, transitions through mid-foot and forefoot contact, and ends with toe-off. The distribution indicates that the proposed interface-enhanced model effectively captures the continuous and spatially distributed contact mechanics of the foot, contributing to more human-like interaction with the ground and enhancing the biomechanical fidelity of the simulation. 

Moreover, Figure.~\ref{fig:contact_points_and_foot_force}(b) presents the corresponding deformation of the foot interface over time. We calculated the strains of the flex edges given the initial shape of the foot, where the positive value represents stretching and the negative value represents contraction. Initial contact (heel strike) shows strong compression in the rearfoot area, consistent with vertical impact absorption as the foot first contacts the ground. Simultaneously, mild stretching appears in the midfoot, indicating distributed load transfer. During midstance, deformation becomes more balanced and diffuse, with smaller-magnitude strain in the midfoot and forefoot. This suggests a more stable support phase where the foot flattens and conforms to the ground surface. As the gait progresses to toe-off, increased tensile strain appears in the forefoot and toe regions, reflecting active pushing and unrolling of the foot. The posterior areas exhibit recovery from earlier compression. These deformation patterns demonstrate that the deformable foot interface dynamically adapts to ground contact phases through distributed and phase-specific structural compliance. This behavior enhances shock absorption, energy return, and terrain adaptability, contributing to more human-like, efficient, and stable locomotion compared to rigid designs. 

According to the stability condition for humans, the Zero Moment Point (ZMP) should remain within the support polygon, defined as the area enclosed by the contact points. Although both the rigid and interface-enhanced models achieve dynamically stable walking, the interface-enhanced model typically exhibits a larger stability margin. The expanded support area contributes to better balance and greater robustness under dynamic conditions, enhancing adaptability to potential perturbations or modeling uncertainties.

\begin{figure}[htbp]
    \centering
    \includegraphics[width=0.8\linewidth]{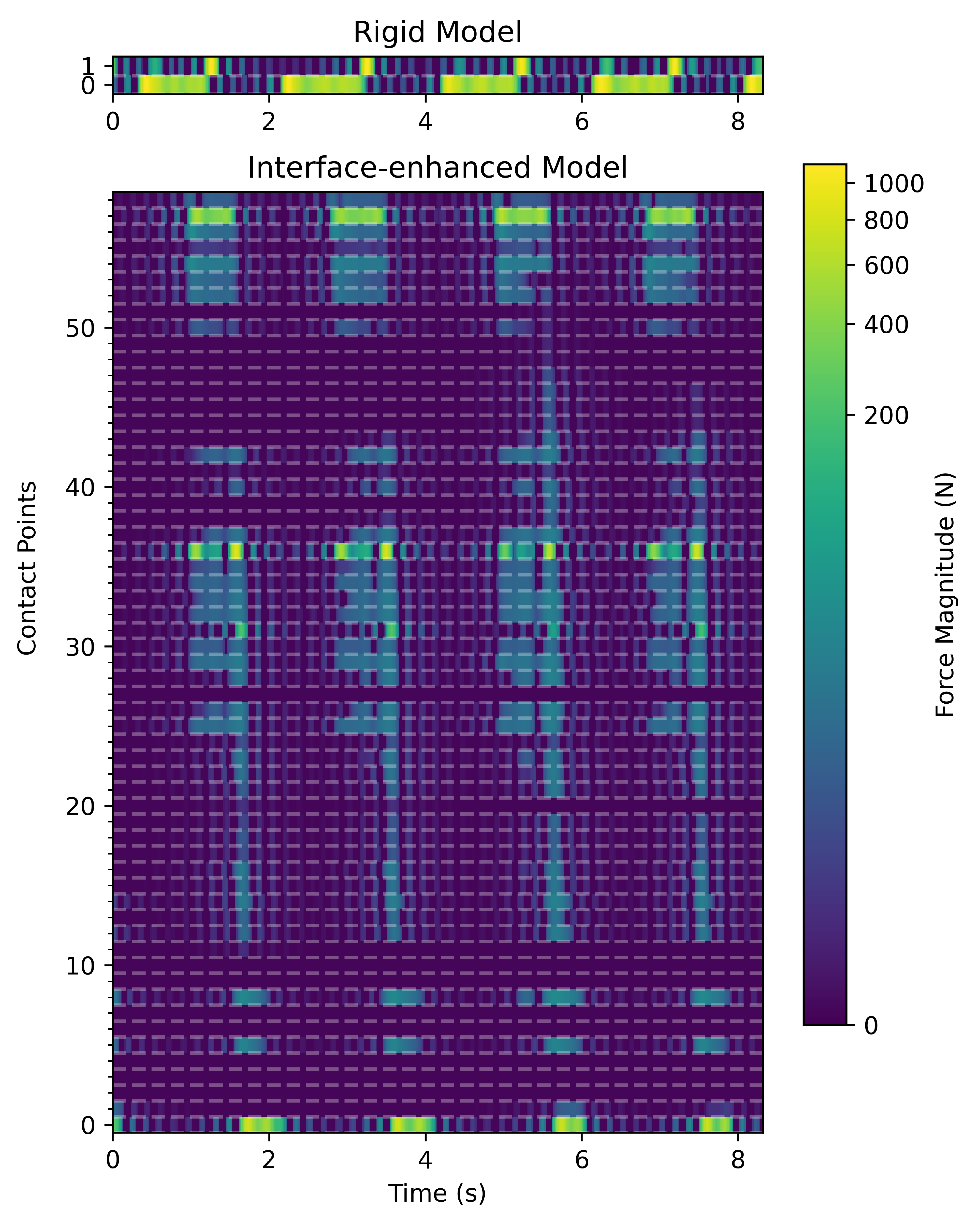}
    \caption{Magnitude of the ground reaction force during walking across contact points in the interface-enhanced and the rigid model. The maximum peak force for the interface-enhanced model and rigid model are 876~$N$ and 1116~$N$ respectively.}
    \label{fig:heatmap}
\end{figure}

\subsubsection{Kinetic Analysis}
\label{sec:kinetic_analysis}

Fig.~\ref{fig:heatmap} compares the ground reaction force (GRF) magnitude over time and contact points for the rigid and interface-enhanced models. The interface-enhanced model engages over 50 contact points and exhibits a broader, more evenly distributed GRF pattern, reflecting the soft and adaptive nature of the deformable interface. This structure spreads the load more uniformly across time and space. However, the rigid model shows a higher peak GRF (1116~$N$) compared to the interface-enhanced model (876~$N$), indicating more concentrated and impulsive force transmission, likely due to its limited compliance and smaller contact area. In terms of temporal characteristics, the rigid model produces sharper, discrete contact events, whereas the interface-enhanced model demonstrates smoother, continuous contact transitions, consistent with a natural foot unrolling during gait phases. These properties enhance stability, comfort, and energy efficiency, and more closely replicate human gait biomechanics than rigid-foot models.

\begin{figure}[tbp]
    \centering
    \includegraphics[width=1\linewidth]{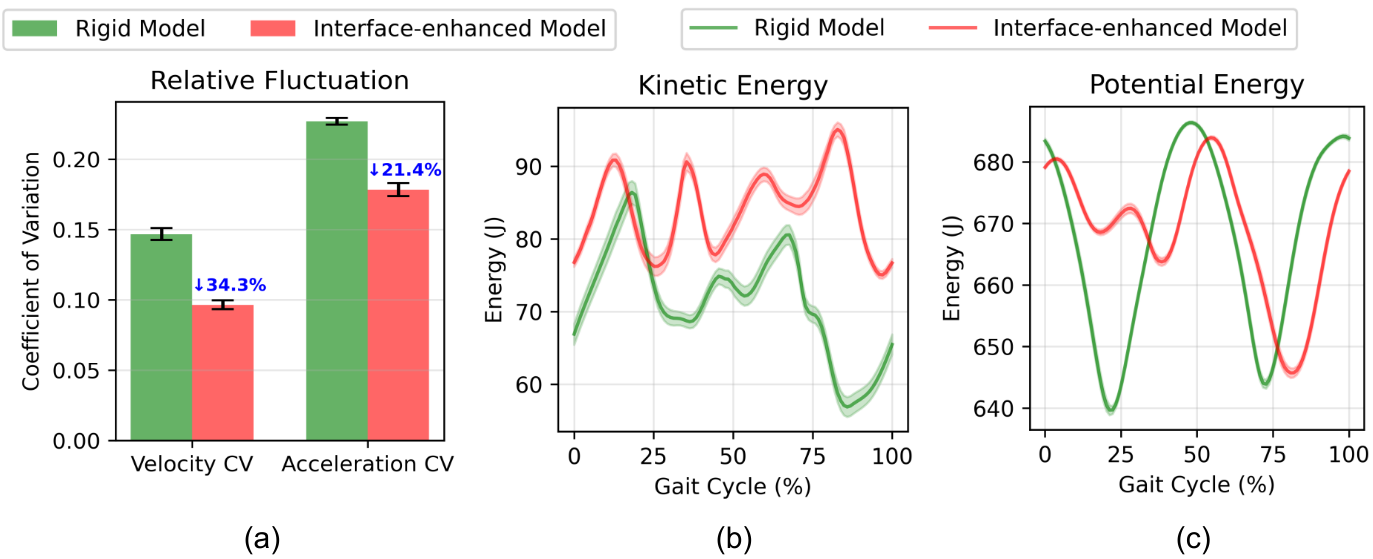}
    \caption{Kinematics and kinetics of the interface-enhanced and the rigid model. (a) Fluctuation of the velocity and acceleration represented by the coefficient of variation (CV). (b) The kinetic energy within one gait cycle. (c) The potential energy within one gait cycle.}
    \label{fig:energy}
\end{figure}

Continuous and smooth force distribution results in lower fluctuations in kinetic performance, as shown in Fig.~\ref{fig:energy}(a). The interface-enhanced model achieves a mean velocity of 1.38~$m/s$ and a mean acceleration of 9.98~$m/s^2$ respectively, comparable to the rigid model's 1.29~$m/s$ and 10.03~$m/s^2$. Although both models exhibit similar velocity and acceleration magnitudes, the interface-enhanced model shows a 34.3$\%$ and 21.4$\%$ reduction in the coefficient of variation (CV) for velocity and acceleration, respectively, indicating more consistent and stable locomotion patterns.


Further analysis of the energy profiles is shown in Fig.~\ref{fig:energy}(b)(c). The interface-enhanced model exhibits consistently higher kinetic energy of 84.0 ± 1.1~$J$ across the gait cycle compared to the rigid model with 71.5 ± 1.3~$J$. Notably, the increase in kinetic energy exceeds the squared increase in velocity, suggesting more dynamic leg motion and greater engagement of the compliant interface. In contrast, the potential energy profile of the interface-enhanced model exhibits smoother, lower-amplitude fluctuations, indicating a more elastic and controlled modulation of the center of mass (COM) elevation. This smoother energy exchange implies that the compliant foot interface supports a more efficient energy storage–return mechanism, analogous to the role of the human foot’s arch and soft tissues during walking. Conversely, the rigid model displays more abrupt potential energy changes and larger variance, reflecting less efficient mechanical performance and reduced capacity for energy reutilization.

Overall, these findings indicate that the compliant interface enhances locomotor efficiency and realism, particularly under conditions demanding stable and adaptive ground interaction. The interface-enhanced musculoskeletal model not only more faithfully reproduces human-like ground contact mechanics, but also confers mechanical advantages in impact attenuation and propulsion efficiency.

\subsection{Real-world Demonstration}
\label{sec:real_world_demonstration}

For the human experiments, insole sensors were employed to record the plantar pressure distribution, while the real-world walking trajectory was captured simultaneously for kinematic and kinetic validation. The subject, shown in Fig.~\ref{fig:whole_pipeline}(a), was a 23-year-old male (70~$kg$) wearing shoes instrumented with insole sensors. Each insole contained 18 sensing channels positioned on the plantar surface to measure vertical ground reaction forces. The recorded force data were temporally aligned with the trajectory sequence and denoised through filtering. Subsequently, the normal force distributions over the forefoot and heel regions were compared between the simulated results and the measured human data.

\begin{figure}[tbp]
    \centering
    \includegraphics[width=1\linewidth]{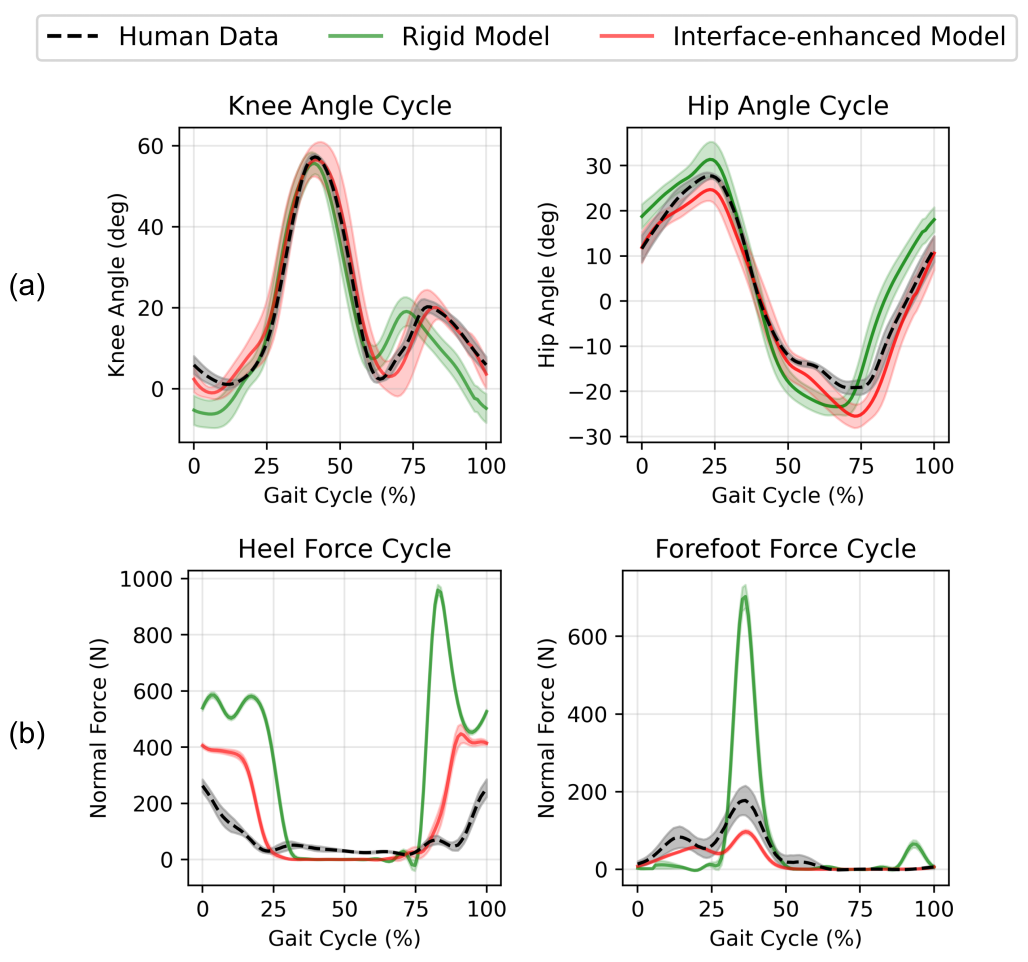}
    \caption{Kinematics and kinetics of the real-world human data (dashed black line), the interface-enhanced musculoskeletal model (solid red line) and the rigid model (solid green line) within one gait cycle. (a) Angle variation of the knee (left) and the hip (right). (b) The normal force distribution at the heel (left) and the forefoot (right).}
    \label{fig:real_world_demonstration}
\end{figure}

As shown in Fig.~\ref{fig:real_world_demonstration}(a), the knee and hip joint angles generated by the interface-enhanced model align more closely with the reference trajectories compared to those of the rigid model. The smaller deviations and narrower confidence intervals suggest that the compliant interface enhances both the accuracy and consistency of kinematic reproduction.
In Fig.~\ref{fig:real_world_demonstration}(b), despite peak discrepancies likely due to sensor drift and limited accuracy, the interface-enhanced model exhibits force distributions that more closely match human data in both timing and magnitude. At the heel, it captures the initial impact phase (0–20$\%$ of the gait cycle) and terminal contact phase (80–100$\%$) with moderate force magnitudes ($\sim$400~$N$), whereas the rigid model exhibits exaggerated peaks exceeding 900~$N$. At the forefoot, the interface-enhanced model aligns better with human measurements and reproduces the double-peaked profile around 15$\%$ and 35$\%$ of the gait cycle, while the rigid model shows a sharp, unrealistic spike near mid-stance. These observations suggest that the compliant interface enables more gradual and physiologically accurate load transfer during foot-ground interaction, mitigating excessive force concentrations characteristic of rigid contacts. The improved correspondence with experimental data validates the physical realism of the interface-enhanced model and underscores its potential for simulating natural locomotion.

\section{CONCLUSIONS}

This study addresses a long-standing limitation in musculoskeletal modeling by integrating a contact-rich, deformable foot model into a conventional rigid musculoskeletal model. Through a two-stage policy training strategy, we achieved natural walking despite the significant computational challenges posed by complex contact dynamics. Comprehensive evaluation revealed that our interface-enhanced model substantially outperforms traditional rigid models across multiple metrics, including: (1) physiologically accurate distribution and continuity of ground reaction forces, (2) significantly improved motion stability in velocity and acceleration variation, and (3) biomechanically efficient energy transfer patterns throughout the gait cycle that closely mirror human locomotion. Validation against human subject data using insole pressure mapping confirmed the high fidelity of our model in reproducing foot-ground interaction dynamics, particularly in terms of joint kinematics and plantar pressure distribution. 

Simulation-based development inevitably limits the realism and accuracy of physical properties and dynamic responses. Advances in physics engines are expected to enhance the fidelity of deformable contact modeling. The lack of comprehensive physiological data limits further validation and clinical applications. Future research directions include adapting this framework for more healthy and pathological gait patterns, and translating these insights into practical applications for exoskeleton design and humanoid robotics, where precise modeling of foot-ground interactions is fundamental to achieving stable, energy-efficient, and human-like locomotion.

\addtolength{\textheight}{-12cm}   









\bibliographystyle{IEEEtran}
\bibliography{IEEEabrv,myref}

\end{document}